\documentclass[11pt]{article}

\usepackage[preprint]{acl}
\usepackage{times}
\usepackage{latexsym}
\usepackage[T1]{fontenc}
\usepackage[utf8]{inputenc}
\usepackage{microtype}
\usepackage{inconsolata}
\usepackage{hyperref}
\usepackage{graphicx}
\usepackage{booktabs}
\usepackage{colortbl}
\usepackage{multirow}
\usepackage{amsmath}
\usepackage{amssymb}
\usepackage{algorithm}
\usepackage{algpseudocode}
\usepackage{xcolor}
\usepackage[most]{tcolorbox}

\definecolor{pbBlue}{HTML}{1F77B4}
\definecolor{pbOrange}{HTML}{E76F00}
\definecolor{pbGreen}{HTML}{2CA02C}
\definecolor{pbRed}{HTML}{C0392B}
\definecolor{pbPurple}{HTML}{8E44AD}
\definecolor{pbTeal}{HTML}{16A085}
\definecolor{pbBrown}{HTML}{8C564B}
\definecolor{pbSlate}{HTML}{34495E}

\newtcolorbox{promptbox}[2][pbBlue]{%
    enhanced,
    breakable,
    colback=#1!4!white,
    colframe=#1!70!black,
    coltitle=white,
    colbacktitle=#1!85!black,
    fonttitle=\bfseries\small,
    title=#2,
    arc=1.2mm,
    boxrule=0.6pt,
    titlerule=0pt,
    left=5pt, right=5pt, top=4pt, bottom=4pt,
    before skip=6pt, after skip=6pt,
    attach boxed title to top left={xshift=4mm,yshift=-2.4mm},
    boxed title style={arc=0.8mm,boxrule=0pt},
    fontupper=\small,
}

\newcommand{\opr}{\textsc{OPR}}
\newcommand{\sdft}{\textsc{SDFT}}
\newcommand{\bwt}{\mathrm{BWT}}
\newcommand{\R}{\mathcal{R}}
\newcommand{\D}{\mathcal{D}}
\newcommand{\B}{\mathcal{B}}

\title{On-Policy Replay for Continual Supervised Fine-Tuning}

\author{
 \textbf{Yan Chen}\textsuperscript{1,$\dagger$}
 \textbf{Taojie Zhu}\textsuperscript{1,$\dagger$},
 \textbf{Meng Zhang}\textsuperscript{1},
 \textbf{Xin Chen}\textsuperscript{2},
 \textbf{Jiaqi Huang}\textsuperscript{1},
 \textbf{Dongyang Xu}\textsuperscript{1},
 \textbf{Yizhi Wang}\textsuperscript{1,$\ddagger$}
\\
\\
 \textsuperscript{1}Tsinghua University
\\
 \textsuperscript{2}Alibaba Group
\\
 \textsuperscript{$\dagger$}Equal contribution. \quad
 \textsuperscript{$\ddagger$}Corresponding author. 
\\
 Email: \{\texttt{yan-chen24}, \texttt{wangyizhi25}\}\texttt{@mails.tsinghua.edu.cn}
}
\begin{document}
\maketitle

\begin{abstract}
Continual supervised fine-tuning (SFT) is the de facto recipe for adapting
large language models (LLMs) to a stream of downstream tasks, but it
suffers from catastrophic forgetting of earlier capabilities. Recent work
shows that on-policy signals---training on the model's own outputs---reduce
forgetting more reliably than off-policy supervision. Existing on-policy
methods route this signal through a new \emph{training objective} (e.g.,
self-distillation losses with a teacher copy), inheriting an extra forward
pass, schedule sensitivity, and stylistic drift from the teacher.
We instead route the on-policy signal through the \emph{training data
source}. Our method, \textbf{On-Policy Replay} (\opr), rolls out the most
recent checkpoint on a small budget of historical prompts, filters the
generations by a task reward, and replays the surviving (prompt, model
response) pairs as ordinary SFT examples. There is no teacher, no auxiliary
loss, and no on-the-fly distillation. Across three 7--8B
instruction-tuned backbones (Qwen2.5-7B-Instruct, Qwen3-8B,
Llama3.1-8B-Instruct) on the TRACE continual-learning benchmark,
\opr~consistently reduces forgetting; on the sharpest stress test
(Qwen2.5-7B-Instruct, Sequential SFT BWT $-13.93$), \opr~lifts BWT to
$-0.65$ at a $10\%$ replay budget and to $-2.29$ at a $1\%$
budget---a $46\%$ reduction in $|\mathrm{BWT}|$ over a tuned
Vanilla Replay baseline, with $42$--$46\%$ reductions observed across
all three backbones. We give a KL-shrinkage interpretation that places
\opr~and prior on-policy distillation methods on a single axis, and we
present a counterintuitive finding that explains why Vanilla Replay is
already a strong baseline: \emph{low-score} replay is uniformly worse
than Vanilla Replay, demonstrating that the active ingredient in \opr~is
the on-policy distribution, not the response quality alone. 
Our code is available at \url{https://github.com/Yancey2024/OnPolicyReplay}.
\end{abstract}

\section{Introduction}
\label{sec:intro}

\begin{figure*}[t]
  \centering
  \includegraphics[width=\textwidth]{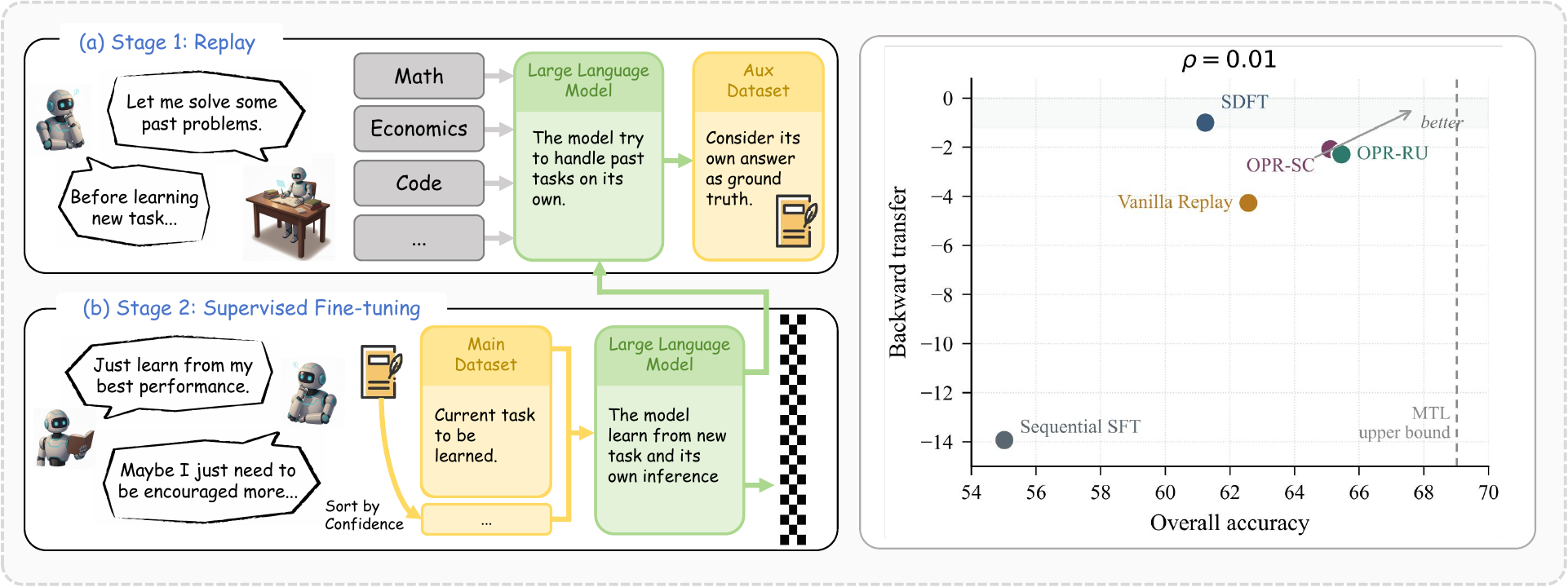}
  \caption{\textbf{On-Policy Replay (\opr) at a glance.} \textit{Left:} after each task, \opr~rolls out on past-task prompts, keeps the top-confidence (prompt, response) pairs as a replay buffer, and mixes it with the next-task data under plain SFT---no teacher, no auxiliary loss. \textit{Right:} BWT vs.~overall accuracy on TRACE/Qwen2.5-7B-Instruct at $\rho{=}0.01$; \opr-RU and \opr-SC move the frontier toward the MTL upper bound.}
  \label{fig:intro}
\end{figure*}

Modern large language models (LLMs) are rarely deployed as one-shot artifacts. A single instruction-tuned checkpoint is repeatedly adapted to new domains, formats, and skills as they emerge after deployment \citep{chung2022scaling,ouyang2022instructgpt,wu2024continual}, and the default tool for each such update is supervised fine-tuning (SFT). Sequential SFT, however, suffers severe \emph{catastrophic forgetting}: accuracy on previously learned tasks degrades by double-digit margins as new tasks are fine-tuned in sequence \citep{luo2023empirical,wang2023trace}. Recent diagnoses converge on a single thread \citep{shenfeld2025rlrazor,chen2025retaining,chu2025sft,zhu2026dypo}: SFT is \emph{off-policy}, and the per-update KL between successive policies on historical prompts, $\mathbb{E}_{x \sim p_{\text{hist}}}\, \mathrm{KL}\!\bigl(\pi_{\theta_t} \,\|\, \pi_{\theta_{t-1}}\bigr)$, upper-bounds how much the model forgets at each step---training on samples from $\pi_{\theta_{t-1}}$ shrinks this KL by construction. The diagnosis sharpens ``mitigate forgetting'' into a concrete question: where in the pipeline should the on-policy signal be delivered?

Existing remedies all answer this question on the \emph{objective} side. \sdft~\citep{shenfeld2026sdft}, OPSD \citep{zhao2026opsd}, and GKD \citep{agarwal2024onpolicy} add an explicit per-token KL term against a teacher copy of the model, paying a teacher forward pass on every gradient step and inheriting a schedule (EMA decay, KL coefficient) that competes with the SFT optimizer for tuning bandwidth. RazorSFT \citep{shenfeld2025rlrazor} keeps plain SFT but replaces the current task's gold target with a self-sampled response that passes a verifier, avoiding the teacher at the price of being confined to the current task's targets. Either way, the on-policy signal lives in the loss or in the target it is computed against. We observe that the underlying constraint is not on the loss but on the \emph{data distribution} the loss sees---and that every replay-based continual learner already has a piece of state with that property: the cross-task replay buffer. The buffer is the one object in the pipeline that carries data across task boundaries; in standard practice it is filled with stale, off-policy gold targets and contributes nothing to the on-policy goal. Making the buffer itself on-policy turns out to be the cheapest way to enforce the KL constraint, and it leaves the training loop untouched.

\begin{figure*}[t]
  \centering
  \includegraphics[width=\textwidth]{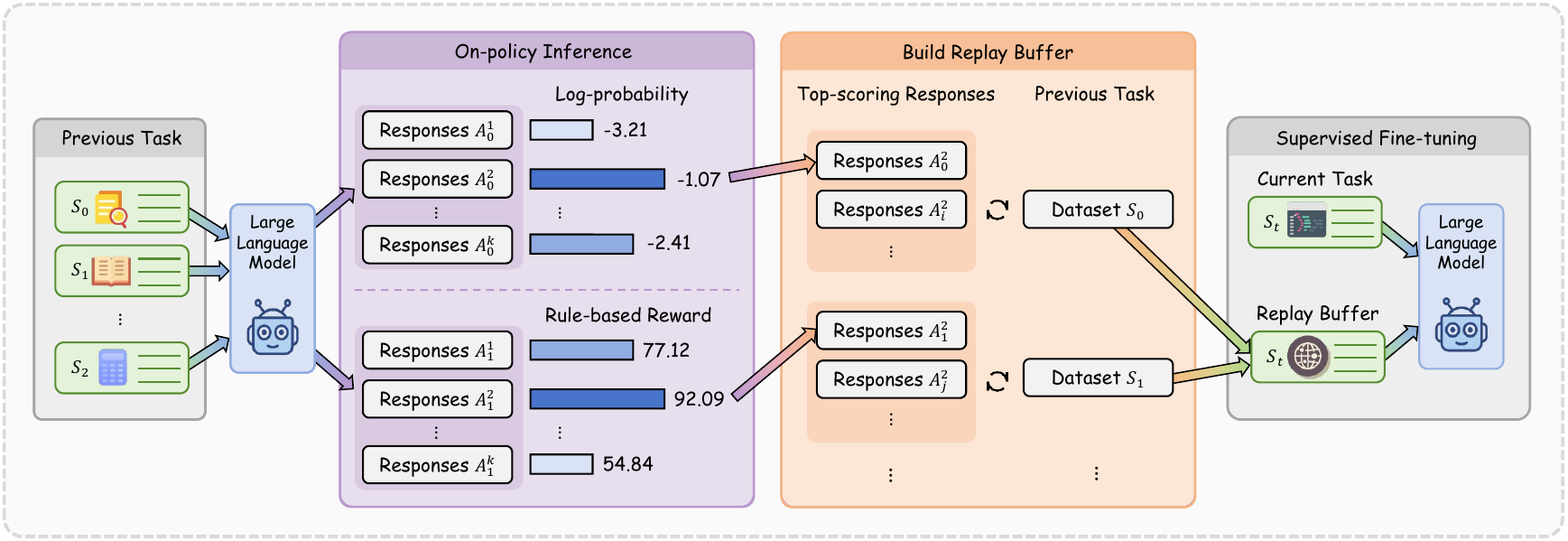}
  \caption{\textbf{\opr~pipeline.} After training on task $T_i$ (dataset $S_i$ in the figure), the checkpoint $\pi_{\theta_i}$ rolls out on historical prompts; a scorer (rule reward or length-normalized log-likelihood) ranks the responses, the top-scoring (prompt, response) pairs form the replay buffer, and the buffer is mixed with task-$T_{i+1}$ data for plain supervised fine-tuning.}
  \label{fig:pipeline}
\end{figure*}

We introduce \textbf{On-Policy Replay} (\opr): after each task, \opr~rolls out the current checkpoint on a small budget of historical prompts, scores each rollout, and keeps only the top-quantile (prompt, response) pairs as the buffer for the next stage of plain cross-entropy SFT---no teacher, no auxiliary loss, no current-target rewrite (Figure~\ref{fig:pipeline}). The default variant, \opr-RU (\emph{rule-based}), scores rollouts with each task's own evaluation metric; because each replayed pair has high $\pi_{\theta_{t-1}}(\tilde{y}\mid x)$ by construction, the SFT gradient on the replay portion is small and points back toward $\pi_{\theta_{t-1}}$, so KL shrinkage emerges as a side effect of an unchanged training loop. We additionally introduce a label-free variant, \opr-SC (\emph{self-confidence}), which scores rollouts by the model's own length-normalized log-probability---returned by the sampler at no additional cost---removing the dependence on task labels for buffer construction.

Across three 7--8B instruction-tuned backbones (Qwen2.5-7B-Instruct~\citep{qwen2025}, Qwen3-8B~\citep{yang2025qwen3}, Llama3.1-8B-Instruct~\citep{grattafiori2024llama3}), \opr~consistently reduces backward transfer. On the sharpest stress test, Qwen2.5-7B-Instruct (Sequential SFT BWT $-13.93$), \opr-RU~lifts BWT to $-2.29$ at a $1\%$ replay budget---a $46\%$ cut in $|\mathrm{BWT}|$ over a tuned Vanilla Replay baseline---and beats \sdft~by $4.2$ overall-accuracy points without a teacher forward pass. A falsifying ablation supports the KL-shrinkage account: replacing the high-reward rollouts in the buffer with low-reward ones from the same pool drives BWT \emph{below} Vanilla Replay, ruling out ``replay-of-anything'' explanations.

\paragraph{Contributions.}
\begin{itemize}
  \itemsep -0.15em
  \item \textbf{\opr}, a continual-SFT method that enforces on-policy KL shrinkage through the cross-task replay buffer rather than through an auxiliary loss term or current-target rewrite, leaving the SFT training loop untouched.
  \item \textbf{Label-free scoring via \opr-SC.} Replacing the rule-based scorer with the model's own length-normalized log-probability of the rollout removes the dependence on ground-truth labels for buffer construction, making continual SFT applicable in settings where reliable task annotations are unavailable.
\end{itemize}

\section{Related Work}
\label{sec:related}

\paragraph{On-policy continual fine-tuning.}
A growing line of work argues that the gradient signal in continual SFT should come from data that lies on the model's own output distribution \citep{chen2025retaining,chu2025sft,zhu2026dypo}, since on-policy data minimizes the per-update KL that bounds forgetting \citep{shenfeld2025rlrazor}. Existing realizations of this principle deliver the signal on the objective side: \sdft~\citep{shenfeld2026sdft}, OPSD \citep{zhao2026opsd}, and GKD \citep{agarwal2024onpolicy} add a per-token KL term against a teacher copy of the model on every gradient step; RazorSFT \citep{shenfeld2025rlrazor} keeps plain SFT but rewrites the current task's gold target with a self-sampled, verifier-passing response. We place a detailed mechanistic comparison in \S\ref{sec:method:kl}; \opr~differs from these methods in that it routes the on-policy signal through the cross-task replay buffer, leaving the SFT loss and target untouched.

\paragraph{Replay-based continual learning for LLMs.}
Among classical anti-forgetting families---regularization \citep{kirkpatrick2017overcoming}, parameter isolation \citep{wang2023orthogonal,ren2024analyzing}, and rehearsal \citep{lopez2017gem}---rehearsal scales most gracefully at LLM scale \citep{wu2024continual,luo2023empirical}, and TRACE \citep{wang2023trace} consolidates eight downstream tasks as a standard testbed. Prior LLM-targeted replay methods refine \emph{which prompts} to anchor on: Vanilla Replay mixes prior tasks' gold data into the current batch \citep{wu2024continual}, GeRe \citep{gere2025} anchors pretraining-time capability with general samples, I-LoRA \citep{ren2024analyzing} replays through interpolated LoRA adapters, the \emph{spurious forgetting} hypothesis \citep{zheng2025spurious} shows that ten anchor instances suffice to recover format alignment, and SuRe \citep{hazard2025sure} selects surprise-prioritised prompts. We refine the orthogonal axis---\emph{which targets} to attach to historical prompts---replacing stale gold answers with reward-filtered on-policy rollouts; \opr-SC additionally removes the dependence on gold labels by scoring rollouts under the model's own length-normalized log-probability.

\section{Method}
\label{sec:method}

\subsection{Preliminaries}
\label{sec:method:setup}

A continual SFT learner observes a sequence of tasks $T_1, \ldots, T_N$, each with a labeled dataset $\D_i = \{(x_{ij}, y_{ij})\}_{j=1}^{n_i}$, and at stage $i$ fine-tunes $\pi_{\theta_{i-1}}$ on $\D_i$ together with an optional replay buffer $\B_i$ of size $|\B_i| \le \rho \cdot n_i$. Let $a_{i,j}$ denote the accuracy of $\pi_{\theta_j}$ on task $T_i$. Backward transfer \citep{lopez2017gem} is $\bwt = \tfrac{1}{N-1}\sum_{i<N}(a_{i,N} - a_{i,i})$, and overall accuracy is the mean of $a_{i,N}$. A negative $\bwt$ measures forgetting.

\subsection{Forgetting as Per-Update KL on Historical Prompts}
\label{sec:method:kl}

We begin from the bound established by \citet{shenfeld2025rlrazor}, which converts ``mitigate forgetting'' into a concrete scalar minimisation problem. For any held-out historical task $T_i$ with prompt distribution $p_i$ and any sufficiently small SFT update from $\theta_{t-1}$ to $\theta_t$ during stage $i$, the cross-entropy loss change on $T_i$ satisfies, to leading order,
\begin{align}
  \Delta_i(\theta_{t-1} \!\to\! \theta_t)
  &\;\le\; C \cdot \mathbb{E}_{x \sim p_i}\!\bigl[\mathrm{KL}(\pi_{\theta_t} \,\|\, \pi_{\theta_{t-1}})\bigr] \nonumber \\
  &\quad + o(\|\theta_t - \theta_{t-1}\|^2),
  \label{eq:kl_bound}
\end{align}
for a constant $C > 0$ that depends on the smoothness of the loss. Aggregating over $j < i$ yields the same bound on the historical pool $p_{\text{hist}} = \tfrac{1}{i-1}\sum_{j < i} p_j$. Forgetting at stage $i$ is therefore upper-bounded by a single scalar---the \emph{per-update KL} between consecutive policies, evaluated on historical prompts---and minimising this scalar minimises the leading-order forgetting term.

The bound is mechanism-agnostic---any algorithm that keeps the per-update KL small mitigates forgetting. We show in \S\ref{sec:method:opr_derivation} that \opr~achieves this as a property of the training data rather than through an explicit loss term: on every replay sample the cross-entropy loss is small at $\theta_{t-1}$ by construction, and the expected SFT gradient over those samples descends $\mathrm{KL}(q_{t-1}\,\|\,\pi_\theta)$, pulling $\pi_\theta$ back toward the previous checkpoint.

\subsection{On-Policy Replay}
\label{sec:method:opr}

\opr~constructs $\B_{i+1}$ from self-generated rollouts of the most recent checkpoint $\pi_{\theta_i}$ on historical prompts (Algorithm~\ref{alg:opr}, Figure~\ref{fig:pipeline}). At the end of stage $i$, $\pi_{\theta_i}$ samples one response per prompt $x \in P_{\le i}$ under low-temperature decoding (sampling details in Appendix~\ref{app:rollout}); a scorer $r$ ranks the resulting (prompt, response) pairs; the top-$b$ entries, with $b = \rho\, n_{i+1}$ distributed equally across the $i$ prior tasks ($b/i$ entries each, with the remainder rounded forward), form $\B_{i+1}$, which is concatenated with $\D_{i+1}$ and trained under the standard token-level cross-entropy. The rollout is amortised once per stage and dominated by training compute in our setting.

\begin{algorithm}[t]
\caption{On-Policy Replay (one stage)}
\label{alg:opr}
\begin{algorithmic}[1]
\Require Checkpoint $\pi_{\theta_i}$; budget $b$; historical prompts $P_{\le i}$; scorer $r$.
\State $\R \leftarrow \emptyset$
\For{each $x \in P_{\le i}$}
  \State $y \sim \pi_{\theta_i}(\cdot \mid x)$;\quad $\R \leftarrow \R \cup \{(x, y, r(x, y))\}$
\EndFor
\State Sort $\R$ by score; $\B_{i+1} \leftarrow$ top $b$ entries.
\State \textbf{Stage} $i{+}1$: train $\pi_{\theta_i}$ on $\D_{i+1} \cup \B_{i+1}$ with cross-entropy SFT.
\State \Return $\pi_{\theta_{i+1}}$
\end{algorithmic}
\end{algorithm}

\subsection{KL-Shrinkage Analysis}
\label{sec:method:opr_derivation}

We now show that the replay portion of every \opr~batch satisfies Eq.~\eqref{eq:kl_bound} \emph{by construction}. Throughout this subsection, $i$ indexes the continual-learning stage and $t$ indexes SGD steps within stage $i+1$; the buffer $\B_{i+1}$ is constructed once at the end of stage $i$ using $\pi_{\theta_i}$, so $\theta_{t-1} = \theta_i$ at the first step of stage $i+1$. Let a buffer item $(x, \tilde{y}) \in \B_{i+1}$ be drawn from the high-reward, on-policy distribution
\begin{equation}
  q_{t-1}(y \mid x) \;=\; \frac{\pi_{\theta_{t-1}}(y \mid x)\,\mathbf{1}[r(x, y) \ge \tau]}{Z_{t-1}(x)},
  \label{eq:q_def}
\end{equation}
where $\tau$ is the per-task top-quantile threshold induced by Algorithm~\ref{alg:opr} and $Z_{t-1}(x)$ is the normaliser. Denote the per-sample cross-entropy loss by $\ell(\theta; x, \tilde{y}) = -\log \pi_\theta(\tilde{y} \mid x)$.

The filter in Algorithm~\ref{alg:opr} acts directly on the loss at $\theta_{t-1}$. For the self-confidence scorer $r_{\text{SC}}$ (\S\ref{sec:method:scorer}, Eq.~\eqref{eq:sc}), the chain rule yields the identity $r_{\text{SC}}(x, \tilde{y}) = -\ell(\theta_{t-1}; x, \tilde{y})/|\tilde{y}|$, so top-$b$ selection at the empirical $b$-quantile $\tau$ is equivalent to a per-sample upper bound on the loss at $\theta_{t-1}$:
\begin{equation}
  \ell(\theta_{t-1}; x, \tilde{y}) \;=\; -\log \pi_{\theta_{t-1}}(\tilde{y} \mid x) \;\le\; \ell_{\max},
  \label{eq:loss_bound}
\end{equation}
where $\ell_{\max} := -|\tilde{y}|\,\tau$; every buffer item sits in a high-likelihood region of $\pi_{\theta_{t-1}}$ by construction. Eq.~\eqref{eq:loss_bound} is tight at the first SGD step of stage $i+1$ (where $\theta_{t-1} = \theta_i$, the policy used to build $\B_{i+1}$) and loosens as $\|\theta_{t-1} - \theta_i\|$ grows, but refreshing the buffer at each stage boundary bounds this drift within a single stage. For the rule-based variant $r_{\text{RU}}$ the identity no longer holds; we adopt the empirical assumption that $r_{\text{RU}}$ is positively correlated with $\log \pi_{\theta_{t-1}}(\tilde{y} \mid x)$ on the rollout pool---task-correct rollouts tend to be high-likelihood ones---so that filtering by $r_{\text{RU}}$ retains rollouts with per-sample loss bounded by an empirical $\ell_{\max}$.

The per-sample bound propagates to the mini-batch gradient. Under $L$-smoothness of $\ell$, the self-bounding inequality $\|\nabla_\theta \ell(\theta_{t-1}; x, \tilde{y})\|^2 \le 2L\,\ell(\theta_{t-1}; x, \tilde{y})$ combined with Jensen's $\|g\|^2 \le \mathbb{E}_{q_{t-1}}[\|\nabla_\theta \ell\|^2]$ and Eq.~\eqref{eq:loss_bound} yields
\begin{equation}
  \|g\|^2 \;\le\; 2L\,\mathbb{E}_{q_{t-1}}[\ell] \;\le\; 2L\,\ell_{\max},
  \label{eq:grad_norm}
\end{equation}
with $g := \mathbb{E}_{q_{t-1}}[\nabla_\theta \ell]$ the gradient on the replay portion of the batch. This gradient also points in a meaningful direction: because $\mathbb{E}_q[\nabla_\theta \log \pi_\theta(y \mid x)] = -\nabla_\theta\,\mathrm{KL}(q \,\|\, \pi_\theta)$ and the entropy of $q$ is independent of $\theta$,
\begin{equation}
  \mathbb{E}_{q_{t-1}}\!\bigl[\nabla_\theta \ell\bigr]
  \;=\;
  \nabla_\theta\,\mathbb{E}_{x}\!\bigl[\mathrm{KL}(q_{t-1} \,\|\, \pi_\theta)\bigr],
  \label{eq:kl_grad}
\end{equation}
so an SGD step against $g$ decreases $\mathrm{KL}(q_{t-1} \,\|\, \pi_\theta)$ and pulls $\pi_\theta$ toward $q_{t-1}$---a restriction of the previous checkpoint to its high-reward region.

Combining these ingredients yields the forgetting bound. Under (a) non-negativity and $L$-smoothness of $\ell$ in $\theta$ on a neighbourhood of $\theta_{t-1}$; (b) bounded spectral norm $\sigma_{\max}$ of the Fisher information $F(\theta_{t-1})$; and (c) step size $\eta$ small enough that $O(\eta^3)$ terms are negligible, one SGD step of size $\eta$ against the replay gradient $g$ satisfies
\begin{equation}
  \mathbb{E}_{p_{\text{hist}}}\!\bigl[\mathrm{KL}(\pi_{\theta_t} \,\|\, \pi_{\theta_{t-1}})\bigr]
  \;\le\;
  L\,\sigma_{\max}\,\eta^2\,\ell_{\max} + o(\eta^2),
  \label{eq:prop1}
\end{equation}
where $\ell_{\max}$ is the per-sample loss bound at $\theta_{t-1}$ enforced by the high-score filter (Eq.~\eqref{eq:loss_bound}). The filter therefore directly controls the leading-order forgetting term. The bound follows from Taylor-expanding $\mathrm{KL}(\pi_{\theta_t} \,\|\, \pi_{\theta_{t-1}})$ in $\theta_t$ around $\theta_t = \theta_{t-1}$, with displacement $\Delta := \theta_t - \theta_{t-1} = -\eta g$. The zeroth-order term is $\mathrm{KL}(\pi_{\theta_{t-1}} \,\|\, \pi_{\theta_{t-1}}) = 0$, and the first-order term vanishes because $\theta_t = \theta_{t-1}$ is the unique minimiser of $\mathrm{KL}(\pi_{\theta_t} \,\|\, \pi_{\theta_{t-1}})$ in $\theta_t$. The leading non-zero term is the quadratic
\begin{align*}
  \tfrac{1}{2}\,\Delta^\top F(\theta_{t-1})\,\Delta
  &= \tfrac{1}{2}\,\eta^2\,g^\top F(\theta_{t-1})\,g \\
  &\le \tfrac{1}{2}\,\sigma_{\max}\,\eta^2\,\|g\|^2,
\end{align*}
where the Hessian of $\mathrm{KL}(\pi_{\theta_t}\|\pi_{\theta_{t-1}})$ in $\theta_t$ at $\theta_t = \theta_{t-1}$ coincides with the Fisher information $F(\theta_{t-1}) = \mathbb{E}\bigl[\nabla\log\pi_{\theta_{t-1}}\,\nabla\log\pi_{\theta_{t-1}}^\top\bigr]$. Substituting Eq.~\eqref{eq:grad_norm} for $\|g\|^2$ yields Eq.~\eqref{eq:prop1}.

The bound in Eq.~\eqref{eq:prop1} rests on \emph{both} ingredients of $q_{t-1}$, which makes a sharp falsifying prediction. Removing high-reward selection while preserving on-policy sampling---i.e., replacing the indicator $\mathbf{1}[r \ge \tau]$ in Eq.~\eqref{eq:q_def} with $\mathbf{1}[r \le \tau]$---keeps Eq.~\eqref{eq:kl_grad} intact but breaks Eq.~\eqref{eq:loss_bound}: low-score rollouts lie in low-likelihood regions of $\pi_{\theta_{t-1}}$, so $\ell_{\max}$ becomes large and Eq.~\eqref{eq:prop1} becomes loose. The theory therefore predicts a degraded $\bwt$ relative to high-score \opr, and predicts that the resulting buffer can be \emph{worse} than naive off-policy replay---since the latter at least uses correct gold targets. We verify both predictions in \S\ref{sec:exp:ablation}.

Loss-side methods such as \sdft~minimise an upper bound on Eq.~\eqref{eq:kl_bound} by adding an explicit term $\beta \cdot \mathrm{KL}(\pi_\theta \,\|\, \pi_\phi)$ to the training loss, where $\pi_\phi$ is a teacher copy of the model. The two routes share the surrogate but encode it differently: as a per-step term in the loss vs.~as a property of the replay data distribution. The data route is intrinsically \emph{one-sided}---Eq.~\eqref{eq:prop1} bounds the per-update KL only on the replay portion of each batch, leaving the new-task gradient unconstrained.

\subsection{Choice of Scorer}
\label{sec:method:scorer}

The scorer $r$ in Algorithm~\ref{alg:opr} is the only component that requires task-level knowledge. We consider two instantiations.

\paragraph{Rule-based (\opr-RU).}
The default variant scores each rollout with the task's own evaluation metric: exact-match for classification (C-STANCE, FOMC, NumGLUE-cm/ds, ScienceQA), token-level fuzz similarity for Py150, ROUGE-L F1 for MeetingBank, SARI for 20Minuten. Because the same scorer is used to filter the buffer and to evaluate the model, the buffer's filter is by construction aligned with the metric the model is judged on.

\paragraph{Self-confidence (\opr-SC).}
Eq.~\eqref{eq:loss_bound} in the derivation above only requires that $\tilde{y}$ lie in a high-likelihood region of $\pi_{\theta_{t-1}}$; the model's own length-normalised log-probability is exactly that quantity:
\begin{equation}
  r_{\text{SC}}(x, y) \;=\; \frac{1}{|y|}\sum_{t=1}^{|y|} \log \pi_{\theta_i}(y_t \mid x, y_{<t}).
  \label{eq:sc}
\end{equation}
Two consequences. First, $r_{\text{SC}}$ is a free by-product of the rollout: the log-probabilities are returned by the sampler at no additional cost. Second, no ground-truth answer, external scorer, or teacher enters the buffer-construction loop---making \opr-SC applicable in production settings where reliable task annotations are unavailable. \S\ref{sec:exp:scoring} reports its head-to-head gap to \opr-RU.

\section{Experiments}
\label{sec:exp}

\begin{table*}[!t]
  \small
  \centering
  \setlength{\tabcolsep}{4pt}
  \begin{tabular}{lcccccccccc}
  \toprule
  \textbf{Method} & \textbf{ACC} & \textbf{BWT}
  & \textbf{C-STA} & \textbf{FOMC} & \textbf{MeBa}
  & \textbf{Py150} & \textbf{SciQA} & \textbf{NG-cm}
  & \textbf{NG-ds} & \textbf{20Min} \\
  \midrule
  \rowcolor{gray!15} \multicolumn{11}{l}{\textit{Qwen2.5-7B-Instruct}} \\
  SFT             & 55.02              & $-13.93$              & 46.23              & 28.41              & 59.23              & 63.09              & 85.74              & 61.64              & \underline{77.15}  & 39.93 \\
  Replay          & 62.57              & $-4.27$               & 53.51              & 69.91              & 59.74              & \underline{64.78}  & \textbf{93.42}     & 65.03              & 77.08              & 40.72 \\
  \sdft           & 61.23              & $\mathbf{-1.00}$      & \underline{54.45}  & 69.43              & 27.78              & 58.30              & \underline{92.80}  & \textbf{76.38}     & 69.19              & \underline{41.35} \\
  \opr-RU         & \textbf{65.45}     & $-2.29$               & \textbf{54.65}     & \textbf{72.65}     & \textbf{63.76}     & 64.07              & 86.56              & \underline{67.67}  & \textbf{77.56}     & \textbf{41.89} \\
  \opr-SC         & \underline{65.11}  & $\underline{-2.08}$   & 51.86              & \underline{70.69}  & \underline{63.60}  & \textbf{66.85}     & 91.68              & 59.57              & 75.31              & 41.33 \\
  \midrule
  \rowcolor{gray!15} \multicolumn{11}{l}{\textit{Qwen3-8B}} \\
  SFT             & 58.25              & $-10.85$              & 46.62              & 28.81              & 59.63              & 63.49              & 86.14              & 62.04              & \underline{77.96}  & 41.33 \\
  Replay          & 65.89              & $-3.37$               & 53.51              & 70.31              & 60.14              & 65.18              & \underline{93.82}  & 65.43              & 77.58              & 41.12 \\
  \sdft           & 64.56              & $\mathbf{-1.20}$      & \underline{54.65}  & 69.83              & 37.18              & 58.70              & 93.20              & \textbf{76.78}     & 69.69              & 41.75 \\
  \opr-RU         & \textbf{68.10}     & $-1.97$               & \textbf{56.53}     & \textbf{74.73}     & \underline{65.84}  & \underline{66.15}  & 88.64              & \underline{69.75}  & \textbf{79.23}     & \underline{42.97} \\
  \opr-SC         & \underline{67.81}  & $\underline{-1.77}$   & 54.44              & \underline{73.27}  & \textbf{66.18}     & \textbf{69.43}     & \textbf{94.26}     & 62.15              & 77.89              & \textbf{43.91} \\
  \midrule
  \rowcolor{gray!15} \multicolumn{11}{l}{\textit{Llama3.1-8B-Instruct}} \\
  SFT             & 50.78              & $-17.07$              & 23.26              & 43.69              & 60.59              & \underline{64.27}  & 72.83              & 28.09              & \underline{72.15}  & 41.32 \\
  Replay          & 58.33              & $-5.36$               & \underline{37.29}  & 69.00              & 61.18              & 64.22              & \underline{82.83}  & 39.82              & 71.15              & 41.21 \\
  \sdft           & 57.01              & $\mathbf{-1.34}$      & \textbf{38.83}     & 70.11              & 30.51              & 59.32              & 80.49              & \underline{41.15}  & 70.84              & \underline{42.72} \\
  \opr-RU         & \textbf{61.89}     & $-2.89$               & 36.90              & \textbf{71.62}     & \textbf{62.53}     & 64.26              & 82.14              & \textbf{41.61}     & \textbf{72.26}     & \textbf{42.82} \\
  \opr-SC         & \underline{61.52}  & $\underline{-2.67}$   & 35.53              & \underline{70.91}  & \underline{62.36}  & \textbf{65.77}     & \textbf{84.50}     & 39.27              & 71.70              & 42.43 \\
  \bottomrule
  \end{tabular}
  \caption{\textbf{Main results on TRACE at $\rho{=}0.01$.} Overall accuracy (ACC, \%), backward transfer (BWT, pp), and per-task accuracy. \textbf{Bold}: best within backbone; \underline{underline}: second-best.}
  \label{tab:main}
\end{table*}

\subsection{Setup}
\label{sec:exp:setup}

\paragraph{Benchmark.}
We conduct our evaluation on TRACE \citep{wang2023trace}, a widely adopted continual learning benchmark for LLMs. TRACE comprises eight diverse tasks arranged in a fixed canonical order: C-STANCE (\emph{C-STA}; stance classification), FOMC (Federal Open Market Committee classification), MeetingBank (\emph{MeBa}; meeting summarization), Py150 (Python code completion), ScienceQA (\emph{SciQA}; multi-step science QA), NumGLUE-cm and NumGLUE-ds (\emph{NG-cm}, \emph{NG-ds}; numerical reasoning), and 20Minuten (\emph{20Min}; German news simplification). For each task, we subsample 5,000 training examples and adopt the official evaluation set.

\paragraph{Models.}
We report the main results on Qwen2.5-7B-Instruct, Qwen3-8B,
and Llama3.1-8B-Instruct.
All three are instruction-tuned checkpoints obtained through RLHF,
and we regard the preservation of instruction-following ability
across the eight-stage sequence as the practically meaningful
objective.

\paragraph{Training.}
All methods use AdamW (learning rate $1\times 10^{-5}$, cosine
schedule, no warmup, weight decay $0$) with a global batch size of
$128$, max sequence length $2048$, and bf16 precision. We run all experiments with three different random seeds and report the averaged results. Per-task epochs
follow $[5,3,7,5,3,5,5,7]$ (chosen to roughly equalize the number of
gradient steps across tasks). The same hyperparameters are applied to
every method and every backbone, isolating the replay-data source as
the only varying factor. See Appendix~\ref{app:train} for the full
infrastructure setup.

\paragraph{Methods compared.}
\textbf{Sequential SFT} (\emph{SFT} in Table~\ref{tab:main}) 
fine-tunes all model parameters on 
tasks one after another with no replay.
\textbf{Multi-Task Learning} (\emph{MTL} in Figure~\ref{fig:intro})
jointly trains on all eight
task datasets in a single run; it has access to data that the
continual learner does not and serves only as an upper bound.
\textbf{Vanilla Replay} (\emph{Replay} in Table~\ref{tab:main}) 
is a standard replay-based sequential fine-tuning
baseline: at each stage it mixes a $\rho$ fraction of each prior task's
gold training data into the current task's batch.
\textbf{Self-Distillation Fine-Tuning} (\emph{SDFT} in Table~\ref{tab:main}) ~\citep{shenfeld2026sdft}
augments the SFT objective with an explicit per-token KL term against a
teacher copy of the model, anchoring the student's output distribution to
the teacher on every gradient step.
\textbf{\opr-RU (ours)} is Algorithm~\ref{alg:opr} with task-specific
rule-based scoring (\S\ref{sec:method:scorer}).
\textbf{\opr-SC (ours)} replaces the rule-based scorer with the
length-normalized log-probability under $\pi_{\theta_{i}}$
(Eq.~\eqref{eq:sc}).

\paragraph{Metrics.}
We report two metrics. The overall accuracy (\emph{ACC} in Table~\ref{tab:main}) is the mean of the per-task accuracies $a_{i,N}$ at the end of the full eight-stage sequence, and reflects final capability. Backward transfer (\emph{BWT} in Table~\ref{tab:main}) is the average per-task accuracy change between end-of-task and end-of-sequence, $\bwt = \tfrac{1}{N{-}1}\sum_{i<N}(a_{i,N} - a_{i,i})$; a more negative value indicates more forgetting.

\subsection{Main Results}
\label{sec:exp:main}

Table~\ref{tab:main} summarizes performance at the strictest $\rho{=}0.01$ budget. The trends are stable across all three backbones.

\textbf{Sequential SFT collapses} on every backbone (BWT $-10.85$ to $-17.07$~pp), consistent with prior reports of forgetting at LLM scale~\citep{wang2023trace,luo2023empirical}. A $1$\% Vanilla Replay budget recovers most of the degradation~\citep{wu2024continual} but leaves a $3$--$5$~pp BWT gap.

\textbf{\opr-RU consistently improves over Vanilla Replay.} It cuts $|\bwt|$ by $42$--$46\%$ across the three backbones at $\rho{=}0.01$, and the pattern holds across all budgets we test (Table~\ref{tab:budget_sweep}).

\begin{figure*}[!t]
  \centering
  \includegraphics[width=0.95\textwidth]{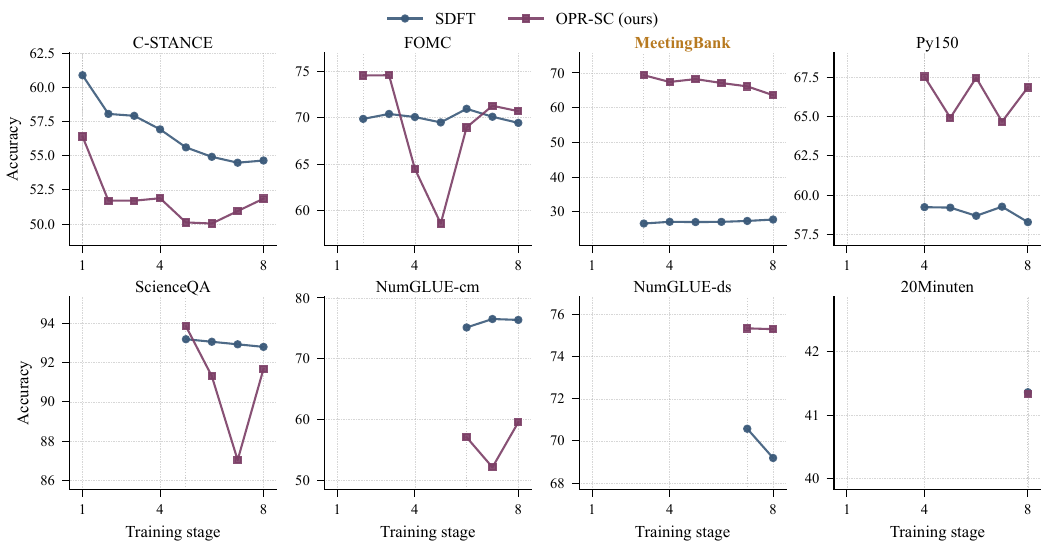}
  \caption{\textbf{Per-task accuracy across the 8 TRACE stages, \sdft~vs.~\opr-SC at $\rho{=}0.01$ on Qwen2.5-7B-Instruct.} \sdft~plateaus on MeetingBank; \opr-SC underperforms on NumGLUE-cm, where the model is confidently wrong.}
  \label{fig:traj_sc}
\end{figure*}

\textbf{\opr~approaches \sdft~without a teacher.} \sdft~retains a marginal BWT edge (within $\sim$$1.5$~pp on every backbone) but loses $3$--$5$~pp of overall accuracy, driven almost entirely by a collapse on MeetingBank (Figure~\ref{fig:traj_sc}). In long-form summarization the self-distilled teacher inherits its own hallucinations; lacking any external reward signal, the student inherits them too. We make this mechanistic in \S\ref{sec:exp:loss}.

\subsection{Ablations Analyses}
\label{sec:exp:ablation}

We sweep three axes: the replay budget $\rho$, a falsifying
low-score variant predicted by the falsifying argument in \S\ref{sec:method:opr_derivation}, and the choice of scorer.

\paragraph{Budget sensitivity.}
We sweep $\rho \in \{0.01, 0.02, 0.05, 0.10\}$ (Table~\ref{tab:budget_sweep}). \opr-RU cuts $|\bwt|$ by $46$--$60\%$ over Vanilla Replay at every budget, and its overall accuracy is nearly flat (range $0.47$~pp) while Vanilla Replay swings by $3.51$~pp---once the buffer is on-policy, replay \emph{quality} dominates over replay \emph{quantity}.

\begin{table}[!t]
  \small
  \centering
  \begin{tabular}{lccc}
  \toprule
  \textbf{Method} & $\boldsymbol{\rho}$ & \textbf{ACC} $\uparrow$ & \textbf{BWT} $\uparrow$ \\
  \midrule
  \multirow{4}{*}{Vanilla Replay}
                  & 0.01 & 62.57          & $-4.27$ \\
                  & 0.02 & 63.68          & $-3.89$ \\
                  & 0.05 & 64.50          & $-3.53$ \\
                  & 0.10 & 66.08          & $-1.37$ \\
  \midrule
  \multirow{4}{*}{\opr-RU (ours)}
                  & 0.01 & 65.45          & $-2.29$ \\
                  & 0.02 & 65.59          & $-1.71$ \\
                  & 0.05 & 65.78          & $-1.43$ \\
                  & 0.10 & \textbf{65.92} & $\mathbf{-0.65}$ \\
  \bottomrule
  \end{tabular}
  \caption{\textbf{Budget sensitivity on Qwen2.5-7B-Instruct.} ACC and BWT for Vanilla Replay vs.~\opr-RU across $\rho \in \{0.01, 0.02, 0.05, 0.10\}$.}
  \label{tab:budget_sweep}
\end{table}

\paragraph{Low-score replay.}
To test the falsifying prediction in \S\ref{sec:method:opr_derivation}, we hold the prompt set, sampler, and code fixed, and swap the high-score rollouts for low-score ones from the same on-policy batch (Table~\ref{tab:ablation_low}, $\rho{=}0.10$). Low-score replay is both (i) much worse than high-score \opr-RU ($\bwt$ degrades $5.3\times$, from $-0.65$ to $-3.47$) and (ii) worse than off-policy Vanilla Replay ($-3.47$ vs.~$-1.37$). On-policy sampling alone is insufficient---the high-reward filter is the decisive factor, exactly as predicted when $\pi_{\theta_{t-1}}(\tilde{y}\mid x)$ is small and the bound in Eq.~\eqref{eq:prop1} becomes loose.

\begin{table}[!t]
  \small
  \centering
  \begin{tabular}{lrr}
  \toprule
  \textbf{Method} ($\rho{=}0.10$) & \textbf{Overall} $\uparrow$ & \textbf{BWT} $\uparrow$ \\
  \midrule
  Vanilla Replay            & 66.08          & $-1.37$ \\
  \opr-RU (high-score)      & \textbf{65.92} & $\mathbf{-0.65}$ \\
  \opr-RU (low-score)       & 65.01          & $-3.47$ \\
  \bottomrule
  \end{tabular}
  \caption{\textbf{Low-score ablation on Qwen2.5-7B-Instruct ($\rho{=}0.10$).} Replacing the high-reward rollouts in the \opr~buffer with low-reward ones from the same pool (prompts, sampler, code held fixed).}
  \label{tab:ablation_low}
  \end{table}

\paragraph{Scorer choice.}
\label{sec:exp:scoring}
At $\rho{=}0.01$ on Qwen2.5-7B-Instruct, \opr-SC matches \opr-RU on aggregate ($65.11$ vs.~$65.45$ ACC; $-2.08$ vs.~$-2.29$ BWT), confirming the design rationale of \S\ref{sec:method:scorer}: the on-policy plus filter structure is the principal driver of the gains, not the specific scorer. The per-task breakdown in Figure~\ref{fig:traj_sc} adds nuance: \opr-SC excels where the model is well-calibrated (FOMC, ScienceQA) but pays a cost on NumGLUE-cm, where fluently-stated incorrect rollouts pass the confidence filter---a known design boundary, since the buffer mechanism anchors what the model is already confident in rather than manufacturing new capability.

\subsection{Loss Dynamics}
\label{sec:exp:loss}

\begin{figure*}[!t]
  \centering
  \includegraphics[width=0.98\textwidth]{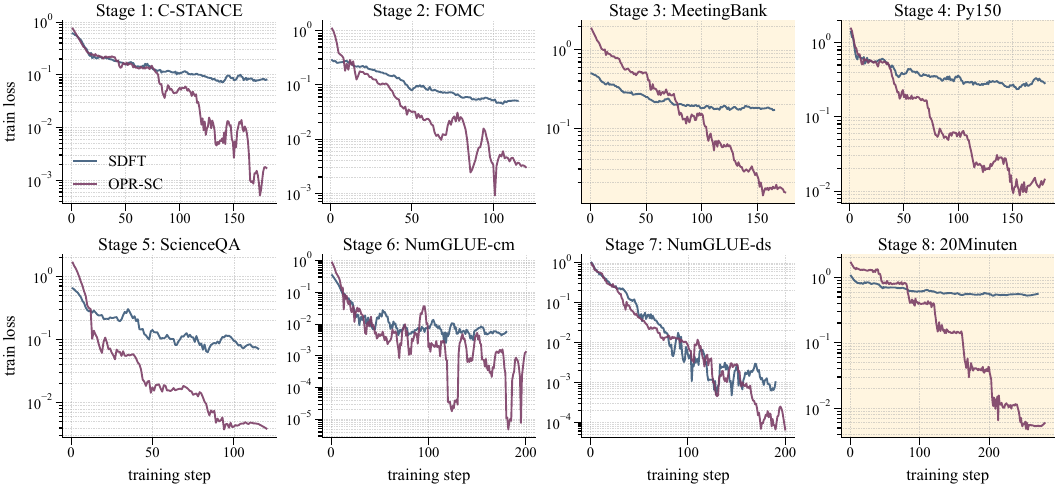}
  \caption{\textbf{Per-stage training loss, \sdft~vs.~\opr-SC at $\rho{=}0.01$ on Qwen2.5-7B-Instruct.} On MeetingBank (Stage 3), Py150 (Stage 4), and 20Minuten (Stage 8) \sdft's loss plateaus while \opr-SC's descends to the same level it reaches on the classification tasks.}
  \label{fig:diag_loss}
\end{figure*}

Figure~\ref{fig:diag_loss} plots per-stage training loss for \sdft~and \opr-SC, sharing every hyperparameter except the supervision route. On four stages (C-STANCE, FOMC, ScienceQA, NumGLUE-ds) the curves are nearly indistinguishable. On three stages---MeetingBank, Py150, and 20Minuten---\sdft's loss plateaus an order of magnitude above \opr-SC's and never descends, while \opr-SC's loss drops to the same low neighbourhood it reaches elsewhere.

The three plateau tasks all demand large distributional shifts from the instruction-tuned base---summarization, code completion, German simplification---whereas the four matched tasks are classification, whose surface form (selecting from labeled options) sits close to the base distribution. \sdft's teacher reverse-KL anchor keeps the cross-entropy from descending precisely on the tasks that need it to descend most, as the bound in \S\ref{sec:method:opr_derivation} predicts.

\subsection{Per-Task Breakdown}
\label{sec:exp:traj}

\begin{figure}[!t]
  \centering
  \includegraphics[width=\columnwidth]{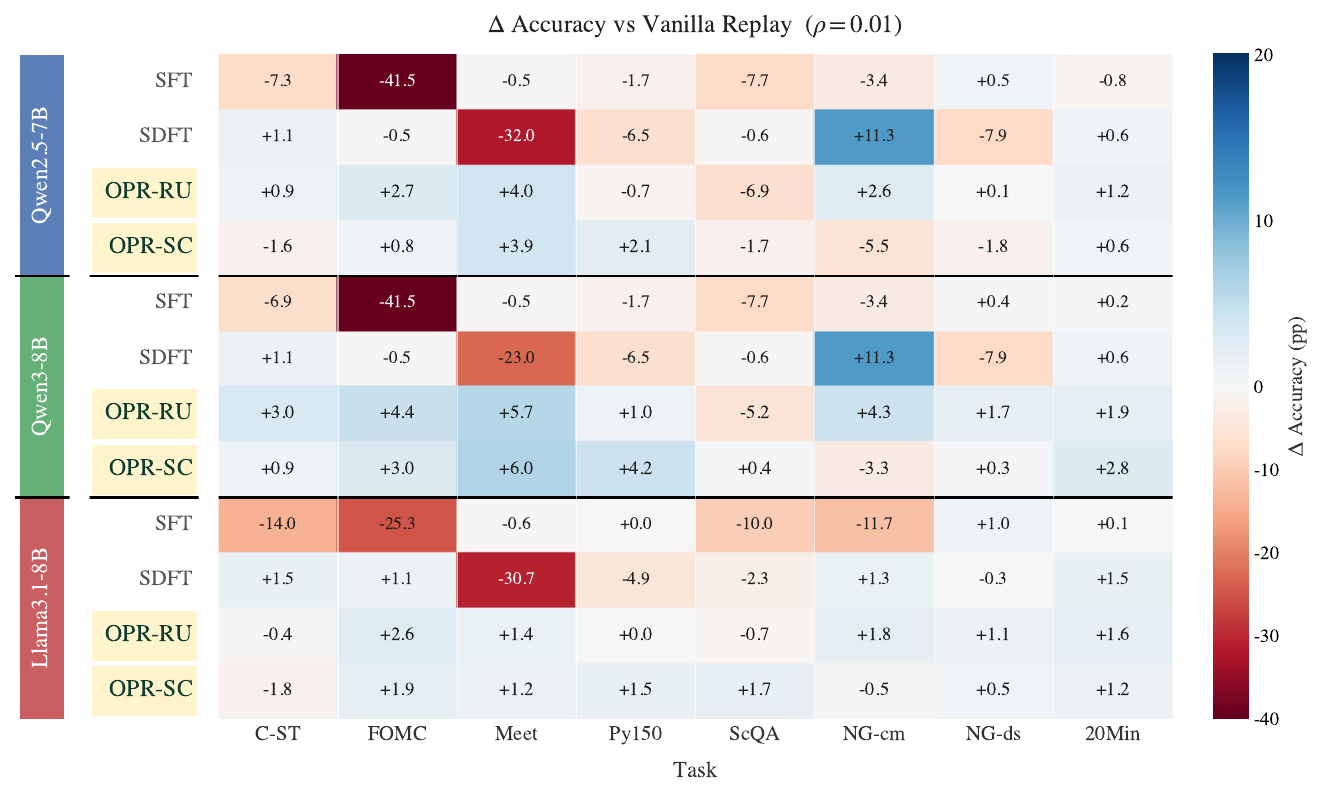}
  \caption{\textbf{Per-task $\Delta$ accuracy vs.~Vanilla Replay at $\rho{=}0.01$.} }
  \label{fig:per_task_delta}
\end{figure}

Figure~\ref{fig:per_task_delta} decomposes Table~\ref{tab:main} into per-task gains and losses relative to Vanilla Replay. \sdft~exhibits a single substantial deficit on MeetingBank ($-23$ to $-32$~pp across backbones), plus smaller consistent deficits on Py150 and NG-ds. Neither \opr-RU nor \opr-SC produces a comparable failure mode---deviations stay within a few points of zero on most tasks. Consistent with \S\ref{sec:exp:loss}: \sdft's anchor pulls the model away from the new-task optimum when a large distributional shift is required, whereas \opr~preserves both retention and acquisition.
\section{Conclusion}
\label{sec:conclusion}

Forgetting in continual SFT is bounded by the per-update KL on historical prompts, and prior work has uniformly enforced that bound on the loss side---via a teacher network, an auxiliary KL term, or a rewritten target. We showed it can be enforced through the cross-task replay buffer alone: drawing the buffer from the previous checkpoint and filtering by reward upper-bounds the per-sample loss at $\theta_{t-1}$ and descends $\mathrm{KL}(q_{t-1}\,\|\,\pi_\theta)$ as an identity, with the SFT loop untouched. Loss-side and data-side methods become two encodings of the same surrogate, and a low-reward ablation falsifies any ``replay-of-anything'' account of the gains. The natural next step is to learn the scorer end-to-end, in place of handcrafting it from evaluation metrics or per-token likelihoods.

\section*{Limitations}

Our claims rest on three $7$--$8$B instruction-tuned backbones and TRACE's canonical eight-task English sequence; we have not validated the KL-shrinkage account on base checkpoints, on models above $\sim$10B parameters, on alternative task orderings, or on non-English continual streams. The headline numbers of \opr-RU also depend on a task-specific evaluator, and although the fallback \opr-SC removes that dependence and matches \opr-RU in aggregate, it fails on a predictable class of inputs---those on which the model is \emph{confidently wrong}---where fluent but incorrect rollouts pass the self-confidence filter and degrade per-task accuracy.

\bibliography{custom}

\appendix
\section{TRACE Tasks, Templates, and Metrics}
\label{app:trace}

This section documents the eight TRACE tasks at the level of detail needed to reproduce our results. It complements the high-level description in
\S\ref{sec:exp:setup}. The original sources of the eight tasks are
C-STANCE~\citep{zhao2023cstance}, FOMC~\citep{shah2023fomc},
MeetingBank~\citep{hu2023meetingbank}, Py150~\citep{raychev2016py150},
ScienceQA~\citep{lu2022scienceqa}, NumGLUE~\citep{mishra2022numglue}
(from which TRACE's NumGLUE-cm and NumGLUE-ds splits are taken), and
20Minuten~\citep{rios202120minuten}.

\subsection{Task Summary}
\label{app:trace:summary}

\begin{table*}[t]
    \small
    \centering
    \begin{tabular}{lccccc}
    \toprule
    \textbf{Dataset} & \textbf{Source} & \textbf{Avg len}
    & \textbf{Metric} & \textbf{Language} & \textbf{Size} \\
    \midrule
    C-STANCE    & Social media & 109  & Accuracy        & Chinese & 5000 \\
    FOMC        & Finance      & 76   & Accuracy        & English & 5000 \\
    MeetingBank & Meeting      & 3550 & ROUGE-L         & English & 5000 \\
    Py150       & Github       & 814  & Edim similarity & English & 5000 \\
    ScienceQA   & Science      & 285  & Accuracy        & English & 5000 \\
    NumGLUE-cm  & Math         & 50   & Accuracy        & English & 5000 \\
    NumGLUE-ds  & Math         & 39   & Accuracy        & English & 5000 \\
    20Minuten   & News         & 739  & SARI            & German  & 5000 \\
    \bottomrule
    \end{tabular}
    
\caption{An overview of the dataset statistics in TRACE. ``Source'' denotes the origin of the context, ``Avg len'' indicates the average number of tokens per dataset, and ``SARI'' refers to a metric specific to text simplification.}
    
    \label{tab:dataset_summary}
    \end{table*}

Two properties of this composition matter for our experiments.
First, average input length varies by almost two orders of
magnitude, from $39$ tokens (NumGLUE-ds) to $3{,}550$ tokens
(MeetingBank), so a per-task replay budget of even $\rho=1\%$
corresponds to very different token counts per task. Second, the
metric set is heterogeneous: ROUGE-L and SARI are bounded surface-form
metrics, edit similarity is character-level, and accuracy is binary,
which means the per-task scores in
Table~\ref{tab:dataset_summary} are not directly commensurable.
This is the same heterogeneity we flag in the Limitations section
and is the reason \opr-RU averages a normalized scalar reward across
tasks rather than the raw metric.

\subsection{Input/Output Examples}
\label{app:trace:examples}

To make the input/output conventions of each TRACE task fully
unambiguous, we provide one representative
\emph{(prompt, gold response)} pair per task below. Each example is
reproduced \emph{verbatim} in the same chat-templated form that is fed
to the model at both training and evaluation time, including
instruction wording, answer-format constraints, and any in-context
formatting (e.g., option letters for multiple-choice tasks). The
examples follow the same order as Table~\ref{tab:dataset_summary}:
C-STANCE, FOMC, MeetingBank, Py150, ScienceQA, NumGLUE-cm,
NumGLUE-ds, and 20Minuten. Note that C-STANCE is natively a Chinese
dataset; the example shown below has been translated into English for
presentation only, while the actual training and evaluation are
performed on the original Chinese text. For readability, long contexts (e.g., the MeetingBank transcripts and 20Minuten articles) are shown in full without truncation, so that the average-length statistics in
Table~\ref{tab:dataset_summary} can be cross-checked against the
samples.

\begin{promptbox}[pbBlue]{Example for C-STANCE}
    \vspace{4pt}
    \textbf{Query: }Determine the stance of the following text toward the given target. Choose one of: A. Support, B. Oppose, C. Neutral. Output only A, B, or C.
    
    Text: Starting bilingual language training for kids from an early age is actually feasible. They can at least handle daily conversations with ease. However, sufficient language input and output must be guaranteed. For example, one parent speaks one language while the other uses the second, and children should respond in the corresponding language. Otherwise, they will end up understanding the language but being unable to speak it.
    
    Target: bilingual parenting for early language training

    \textbf{Answer: }A
    \end{promptbox}

\begin{promptbox}[pbOrange]{Example for FOMC}
    \vspace{4pt}
    \textbf{Query: }What is the monetary policy stance for the following text? A. dovish, B. hawkish, C. neutral. Choose one from A, B and C.

    Text: First, tightening will in all likelihood occur in the context of a more firmly established economic recovery in the United States so that any adverse effects on EME financial conditions should be buffered by the beneficial effects of higher external demand.

    \textbf{Answer: }B
    \end{promptbox}

\begin{promptbox}[pbGreen]{Example for MeetingBank}
    \vspace{4pt}
    \textbf{Query: }Write a summary of the following meeting transcripts.
    
    Meeting transcripts:Councilmember Clark, would you please put resolution 2a1 on the floor for adoption.As council president of the council, resolution 201 be adopted.Thank you. We've got a mover in a second. Comments and questions by members of Council on Resolution 201. Councilmember CdeBaca.Thank you. I just want to go on record voting no for this tonight. Pepsi is not necessarily a group that needs \$1,000,000 incentive to stay in Denver, and they are well on their way to completing their move with or without this incentive.So definitely just want to make sure that I'm clear in not supporting this tonight. Thank you.Thank you. And I'll go ahead. And since this is in my council district, District 11, we as community members, along with some of the original steering committee members from the far northeast neighborhood plan, reconvened and talked through all of the pros and possible challenges of welcoming a large scale manufacturer into District 11.And the community saw fit that this was a great use for this area and this will bring close to 500 jobs to the region as well. And I would ask my colleagues to support it tonight. Madam Secretary, roll call on Council Resolution 22, Dash 201, please.See tobacco? No. Clark.I.Swen.I. Herndon, I.Hi. All right. Cashman. Kenny Ortega. Sandoval.I.Sawyer. I. Torres. I. Black. Madam President, I.Madam Secretary, please close the voting and announce the results.One day, 11 eyes.11 Eyes. Council Resolution 22, dash 201 has been adopted. Madam Secretary, please put the next item up on our screens. Councilmember Flynn, go ahead with your comments on resolution 159, please.

    \textbf{Answer: }A resolution approving a proposed Business Incentive Fund (BIF) contract with Bottling Group, LLC for an amount not to exceed \$1,000,000 for 7 years and for performance based criteria that include establishment of a state-of-the-art manufacturing facility and associated significant investments, as well as job retention and creation and associated investments into upskilling and training of the manufacturing facility's Denver-based workforce.Approves a Business Incentive Fund (BIF) contract with Bottling Group, LLC for an amount not to exceed \$1,000,000 and for 7 years through 12-31-2028 for performance based criteria that include establishment of a state-of-the-art manufacturing facility and associated significant investments, as well as job retention and creation and associated investments into upskilling and training of the manufacturing facility's Denver-based workforce (OEDEV-202160849-00). The last regularly scheduled Council meeting within the 30-day review period is on 4-11-2022. The Committee approved filing this item at its meeting on 3-9-2022.
    \end{promptbox}

\begin{promptbox}[pbRed]{Example for Py150}
    \vspace{4pt}
    \textbf{Query: }code

    \textbf{Answer: }completion
    \end{promptbox}

\begin{promptbox}[pbPurple]{Example for ScienceQA}
    \vspace{4pt}
    \textbf{Query: }Choose an answer for the following question and give your reasons.
    
    Question: What information supports the conclusion that Trent acquired this trait?
    
    Hint: Read the description of a trait. Trent knows how to fly a kite.

    Choices: A. Trent's neighbor taught him how to repair a kite. B. Trent likes to fly a kite with his younger brother. C. Trent's friend taught him how to fly a kite.

    \textbf{Answer: }C. Organisms, including people, have both inherited and acquired traits.
    
    Inherited and acquired traits are gained in different ways. Inherited traits are passed down from biological parents to their offspring through genes. Genes are pieces of hereditary material that contain the instructions that affect inherited traits. Offspring receive their genes, and therefore gain their inherited traits, from their biological parents. Inherited traits do not need to be learned.
    
    Acquired traits are gained during a person's life. Some acquired traits, such as riding a bicycle, are gained by learning. Other acquired traits, such as scars, are caused by the environment. Parents do not pass acquired traits down to their offspring.
    \end{promptbox}

\begin{promptbox}[pbTeal]{Example for NumGLUE-cm}
    \vspace{4pt}
    \textbf{Query: }Solve the following math problem.
    
    Question: In a car parking lot, Greyson observed that there are 48 car wheels. Find out the number of cars in the parking lot.

    \textbf{Answer: }12
    \end{promptbox}

\begin{promptbox}[pbBrown]{Example for NumGLUE-ds}
    \vspace{4pt}
    \textbf{Query: }Solve the following math problem.
    
    Question: How many moles of Ammonium iodide are required to react with 3 moles of Potassium hydroxide to form 3 moles of Ammonia, 3 moles of Potassium iodide and 3 moles of Water. 

    \textbf{Answer: }3
    \end{promptbox}

\begin{promptbox}[pbSlate]{Example for 20Minuten}
    \vspace{4pt}
    \textbf{Query: }Provide a simplified version of the following paragraph in German.
    
    Paragraph: In Lyss (Bern), police raided an illegal gambling den and arrested two women aged 30 and 41. Around 40 people were checked on Friday evening on suspicion of illegal gambling and betting. According to the Bernese Jura-Seeland Regional Public Prosecutor's Office and Bern Cantonal Police, the two women are believed to have worked in the premises without a valid work permit and will be charged with violations of the Foreign Nationals and Integration Act. Authorities confirmed that prohibited gambling and betting services were offered at the site. During the inspection and a subsequent house search, officers seized various electronic devices, a slot machine, tens of thousands of Swiss francs in cash and suspected gold items. The operation followed extensive prior investigations. Public prosecutors are still looking into the operators of the premises. Illegal gambling and sports betting raids have taken place in Lyss multiple times before. Back in 2018, 25 people were checked at another venue, with cash, betting computers and slot machines confiscated. Similar operations were also carried out in 2016 and 2015.
    
    \textbf{Answer: }In Lyss, the Bern Cantonal Police conducted a targeted inspection. Cash was seized and two women were arrested during the operation. Similar inspections had already taken place in previous years.
    \end{promptbox}

\section{Training Details}
\label{app:train}

All runs use the \texttt{swift sft} entry point of ms-swift with
full-parameter fine-tuning, on 8$\times$H100 (80\,GB) GPUs under
DeepSpeed ZeRO-2 and Liger Kernel, in bf16. We apply AdamW with
learning rate $1\times10^{-5}$, cosine schedule, no warmup, weight
decay $0$, per-device batch size $16$ (global $128$), and max
sequence length $2048$. Per-task epochs follow $[5,3,7,5,3,5,5,7]$.
Only the final checkpoint of each stage is retained and used as the
initialization of the next stage. The same hyperparameters are
applied to every method and every backbone, so that the replay-data
source is the only varying factor.The total training cost for a single model across all 8 datasets amounts to 32 GPU-hours.

\section{Rollout for OPR}
\label{app:rollout}

Between consecutive stages, the most recent checkpoint
$\pi_{\theta_i}$ generates one response per historical prompt using
\textbf{vLLM}. The tensor parallel size is $4$, sampling temperature
$\tau=0.1$, and nucleus sampling parameter top-$p$ at its default
$1$, with a maximum of $512$ newly generated tokens. The rollout
runs as a separate subprocess from the training step, which uses
8-way data parallelism, so the rollout GPUs need not hold the
optimizer state.

For each prior task $T_j$, the prompts are identical to the $5{,}000$
training prompts used at stage $j$, with the gold labels
\emph{discarded}. Prompts whose chat-templated length exceeds
$2048$ tokens are removed by an overlong filter, and the same filter
is consistently applied during both training and evaluation.

Given a global budget $b = \rho \cdot 5000$, the buffer at stage
$i{+}1$ is divided among the $i$ prior tasks as evenly as possible,
subject to the constraint that the allocation for any single task
does not exceed the size of its corresponding source buffer.

\section{Detailed Experiment Results}
\label{app:exp_res}

In this section, we report a portion of the detailed experimental results from our paper. The model employed is Qwen2.5-7B-Instruct, and the corresponding results are presented in the following table.

\begin{table*}[t]
    \small
    \centering
    \begin{tabular}{lcccccccc}
    \toprule
    \textbf{Task ID} & \textbf{C-STA} & \textbf{FOMC} & \textbf{MeBa} 
    & \textbf{Py150} & \textbf{SciQA} & \textbf{NG-cm} & \textbf{NG-ds} & \textbf{20Min} \\
    \midrule
    1 & 55.89 &  &  &  &  &  &  &  \\
    2 & 54.37 & 73.69 &  &  &  &  &  &  \\
    3 & 49.88 & 65.80 & 68.30 &  &  &  &  &  \\
    4 & 47.70 & 33.72 & 42.26 & 66.80 &  &  &  &  \\
    5 & 17.40 & 0.00 & 51.93 & 62.83 & 95.09 &  &  &  \\
    6 & 50.74 & 65.78 & 47.89 & 66.27 & 92.40 & 62.96 &  &  \\
    7 & 50.95 & 67.09 & 44.98 & 63.19 & 80.41 & 63.58 & 77.92 &  \\
    8 & 46.23 & 28.41 & 59.23 & 63.09 & 85.74 & 61.64 & 77.15 & 39.93 \\
    \bottomrule
    \end{tabular}
    \caption{Detailed results of Qwen2.5-7B-Instruct under sequential SFT.}
\end{table*}

\begin{table*}[t]
    \small
    \centering
    \begin{tabular}{lcccccccc}
    \toprule
    \textbf{Task ID} & \textbf{C-STA} & \textbf{FOMC} & \textbf{MeBa} 
    & \textbf{Py150} & \textbf{SciQA} & \textbf{NG-cm} & \textbf{NG-ds} & \textbf{20Min} \\
    \midrule
    1 & 55.82 &       &       &       &       &       &       &       \\
    2 & 54.51 & 73.54 &       &       &       &       &       &       \\
    3 & 50.93 & 69.76 & 69.04 &       &       &       &       &       \\
    4 & 50.92 & 69.25 & 58.29 & 67.02 &       &       &       &       \\
    5 & 51.36 & 65.50 & 61.92 & 65.10 & 94.99 &       &       &       \\
    6 & 52.64 & 68.78 & 60.55 & 66.30 & 92.72 & 71.60 &       &       \\
    7 & 52.74 & 68.20 & 60.09 & 66.28 & 92.12 & 65.28 & 77.54 &       \\
    8 & 53.51 & 69.91 & 59.74 & 64.78 & 93.42 & 65.03 & 77.08 & 40.72 \\
    \bottomrule
    \end{tabular}
    \caption{Detailed results of Qwen2.5-7B-Instruct under vanilla replay with $\rho = 0.01$.}
\end{table*}

\begin{table*}[t]
    \small
    \centering
    \begin{tabular}{lcccccccc}
    \toprule
    \textbf{Task ID} & \textbf{C-STA} & \textbf{FOMC} & \textbf{MeBa} 
    & \textbf{Py150} & \textbf{SciQA} & \textbf{NG-cm} & \textbf{NG-ds} & \textbf{20Min} \\
    \midrule
    1 & 60.89 &       &       &       &       &       &       &       \\
    2 & 58.06 & 69.86 &       &       &       &       &       &       \\
    3 & 57.92 & 70.39 & 26.63 &       &       &       &       &       \\
    4 & 56.93 & 70.06 & 27.12 & 59.25 &       &       &       &       \\
    5 & 55.61 & 69.48 & 27.05 & 59.22 & 93.19 &       &       &       \\
    6 & 54.93 & 70.94 & 27.11 & 58.70 & 93.06 & 75.15 &       &       \\
    7 & 54.49 & 70.09 & 27.37 & 59.28 & 92.93 & 76.54 & 70.58 &       \\
    8 & 54.65 & 69.43 & 27.78 & 58.30 & 92.80 & 76.39 & 69.19 & 41.36 \\
    \bottomrule
    \end{tabular}
    \caption{Detailed results of Qwen2.5-7B-Instruct using SDFT}
\end{table*}

\begin{table*}[t]
    \small
    \centering
    \begin{tabular}{lcccccccc}
    \toprule
    \textbf{Task ID} & \textbf{C-STA} & \textbf{FOMC} & \textbf{MeBa} 
    & \textbf{Py150} & \textbf{SciQA} & \textbf{NG-cm} & \textbf{NG-ds} & \textbf{20Min} \\
    \midrule
    1 & 55.79 &       &       &       &       &       &       &       \\
    2 & 53.71 & 74.72 &       &       &       &       &       &       \\
    3 & 50.43 & 71.12 & 69.34 &       &       &       &       &       \\
    4 & 52.25 & 66.73 & 59.38 & 66.03 &       &       &       &       \\
    5 & 52.06 & 63.26 & 61.29 & 63.94 & 94.60 &       &       &       \\
    6 & 52.31 & 70.36 & 64.18 & 65.87 & 93.18 & 69.29 &       &       \\
    7 & 54.46 & 74.32 & 63.20 & 62.30 & 82.01 & 66.20 & 77.96 &       \\
    8 & 54.65 & 72.65 & 63.76 & 64.07 & 86.56 & 67.67 & 77.56 & 41.89 \\
    \bottomrule
    \end{tabular}
    \caption{Detailed results of Qwen2.5-7B-Instruct under OPR-RU with $\rho = 0.01$.}
\end{table*}


\begin{table*}[t]
    \small
    \centering
    \begin{tabular}{lcccccccc}
    \toprule
    \textbf{Task ID} & \textbf{C-STA} & \textbf{FOMC} & \textbf{MeBa} 
    & \textbf{Py150} & \textbf{SciQA} & \textbf{NG-cm} & \textbf{NG-ds} & \textbf{20Min} \\
    \midrule
    1 & 56.41 &       &       &       &       &       &       &       \\
    2 & 51.71 & 74.52 &       &       &       &       &       &       \\
    3 & 51.71 & 74.55 & 69.35 &       &       &       &       &       \\
    4 & 51.89 & 64.49 & 67.41 & 67.53 &       &       &       &       \\
    5 & 50.12 & 58.62 & 68.24 & 64.92 & 93.85 &       &       &       \\
    6 & 50.04 & 68.95 & 67.11 & 67.46 & 91.31 & 57.10 &       &       \\
    7 & 50.95 & 71.27 & 66.14 & 64.67 & 87.04 & 52.16 & 75.35 &       \\
    8 & 51.86 & 70.69 & 63.60 & 66.85 & 91.68 & 59.57 & 75.31 & 41.33 \\
    \bottomrule
    \end{tabular}
    \caption{Detailed results of Qwen2.5-7B-Instruct under OPR-SC with $\rho = 0.01$.}
\end{table*}

\end{document}